\documentclass[10pt,twocolumn,letterpaper]{article}

\usepackage{cvpr}
\usepackage{times}
\usepackage{epsfig}
\usepackage{xspace}
\usepackage{color}
\usepackage{multirow}
\usepackage{graphics}
\usepackage{adjustbox}
\usepackage{subfigure}
\usepackage{amsmath}
\usepackage[noend]{algpseudocode}
\usepackage{algorithmicx,algorithm}
\usepackage{paralist} 
\usepackage{enumitem}
\usepackage{graphicx} 
\usepackage{epstopdf}
\usepackage{booktabs} 

\long\def\ignorethis#1{}

\definecolor{Gray}{rgb}{0.35,0.35,0.35}
\definecolor{Blue}{rgb}{0,0.2,0.8}
\definecolor{Red}{rgb}{0.8,0.2,0}
\definecolor{Green}{rgb}{0.0,0.5,0.1}
\definecolor{Gray}{rgb}{0.4,0.4,0.4}

\newlength\paramargin
\newlength\figmargin
\newlength\secmargin

\usepackage{array}
\newcolumntype{L}[1]{>{\raggedright\let\newline\\\arraybackslash\hspace{0pt}}m{#1}}
\newcolumntype{C}[1]{>{\centering\let\newline\\\arraybackslash\hspace{0pt}}m{#1}}
\newcolumntype{R}[1]{>{\raggedleft\let\newline\\\arraybackslash\hspace{0pt}}m{#1}}

\graphicspath{{figure/eps/}}

\usepackage{times}
\usepackage{epsfig}
\usepackage{graphicx}
\usepackage{amsmath}
\usepackage{amssymb}
\usepackage{booktabs}
\usepackage{subfigure}
\usepackage{multirow}
\definecolor{citecolor}{RGB}{34,139,34}
\usepackage[pagebackref=true,breaklinks=true,letterpaper=true,colorlinks, citecolor=citecolor,bookmarks=false]{hyperref}

\graphicspath{ {figures/} }

\cvprfinalcopy 

\def\cvprPaperID{632} 

\newcommand{\myparagraph}[1]{{\vspace{0.5em} \noindent \bf #1}}

\begin{document}


\title{CrowdHuman: A Benchmark for Detecting Human in a Crowd} 

\author{
    Shuai Shao\thanks{Equal contribution.}
    \hspace{10pt}
    Zijian Zhao\footnotemark[1]
    \hspace{10pt}
    Boxun Li
    \hspace{10pt}
    Tete Xiao
    \hspace{10pt}
    Gang Yu
    \hspace{10pt}
    Xiangyu Zhang
    \hspace{10pt}
    Jian Sun\\
    Megvii Inc. (Face++)\\
{\tt\small \{shaoshuai, zhaozijian, liboxun, xtt, yugang, zhangxiangyu, sunjian\}@megvii.com}
}

\maketitle

\begin{abstract}
Human detection has witnessed impressive progress in recent years. However, the occlusion issue of detecting human in highly crowded environments is far from solved. To make matters worse, crowd scenarios are still under-represented in current human detection benchmarks. In this paper, we introduce a new dataset, called CrowdHuman\footnote{Our CrowdHuman dataset can be downloaded from~\url{https://sshao0516.github.io/CrowdHuman/}}, to better evaluate detectors in crowd scenarios. The CrowdHuman dataset is large, rich-annotated and contains high diversity. There are a total of $470K$ human instances from the train and validation subsets, and $~22.6$ persons per image, with various kinds of occlusions in the dataset. Each human instance is annotated with a head bounding-box, human visible-region bounding-box and human full-body bounding-box. Baseline performance of state-of-the-art detection frameworks on CrowdHuman is presented. The cross-dataset generalization results of CrowdHuman dataset demonstrate state-of-the-art performance on previous dataset including Caltech-USA, CityPersons, and Brainwash without bells and whistles. We hope our dataset will serve as a solid baseline and help promote future research in human detection tasks.
\end{abstract}

\section{Introduction}
Detecting people in images is among the most important components of computer vision and has attracted increasing attention in recent years~\cite{zhang2016faster,li2015scale,zhang2015filtered,zhang2016far,hosang2015taking,dollar2009integral,dollar2014fast,dollar2009pedestrian,mao2017can}. 
A system that is able to detect human accurately plays an essential role in applications such as autonomous cars, smart surveillance, robotics, and advanced human machine interactions. Besides, it is a fundamental component for research topics like multiple-object tracking~\cite{mot2015}, human pose estimation~\cite{yilun2018cpn}, and person search~\cite{xiaotong2017cvpr}. Coupled with the development and blooming of convolutional neural networks (CNNs)~\cite{krizhevsky2012imagenet,simonyan2014very,he2016deep}, modern human detectors~\cite{cai2016unified,zhang2016faster,wang2018Repulsion} have achieved remarkable performance on several major human detection benchmarks. 

However, as the algorithms improve, more challenging datasets are necessary to evaluate human detection systems in more complicated real world scenarios, where crowd scenes are relatively common. In crowd scenarios, different people occlude with each other with high overlaps and cause great difficulty of \emph{crowd occlusion}. For example, when a target pedestrian T is largely overlapped with other pedestrians,  the detector may fail to identify the boundaries of each person as they have similar appearances. Therefore,  detector will treat the crowd as a whole, or shift the target bounding box of $T$ to other pedestrians mistakenly. To make matters worse, even though the detectors are able to discriminate different pedestrians in the crowd, the highly overlapped bounding boxes will also be suppressed by the post process of non-maximum suppression (NMS). As a result,  crowd occlusion makes the detector sensitive to the threshold of NMS. A lower threshold may lead to drastically drop on recall, while a higher threshold brings more false positives.


Current datasets and benchmarks for human detection, such as Caltech-USA~\cite{dollar2009pedestrian}, KITTI~\cite{Geiger2012CVPR}, CityPersons~\cite{zhang2017citypersons}, and ``person'' subset of MSCOCO~\cite{lin2014microsoft}, have contributed to a rapid progress in the human detection. Nevertheless, crowd scenarios are still under-represented in these datasets. For example, the statistical number of persons per image is only $0.32$ in Caltech-USA, $4.01$ in COCOPersons, and $6.47$ in CityPersons.  And the average of pairwise overlap between two human instances (larger than 0.5 IoU) in these datasets is only $0.02$, $0.02$, and $0.32$, respectively. Furthermore, the annotators for these datasets are more likely to annotate crowd human as a whole ignored region, which cannot be counted as valid samples in training and evaluation.

\begin{figure*}[thp]
\centering
\includegraphics[width=17cm]{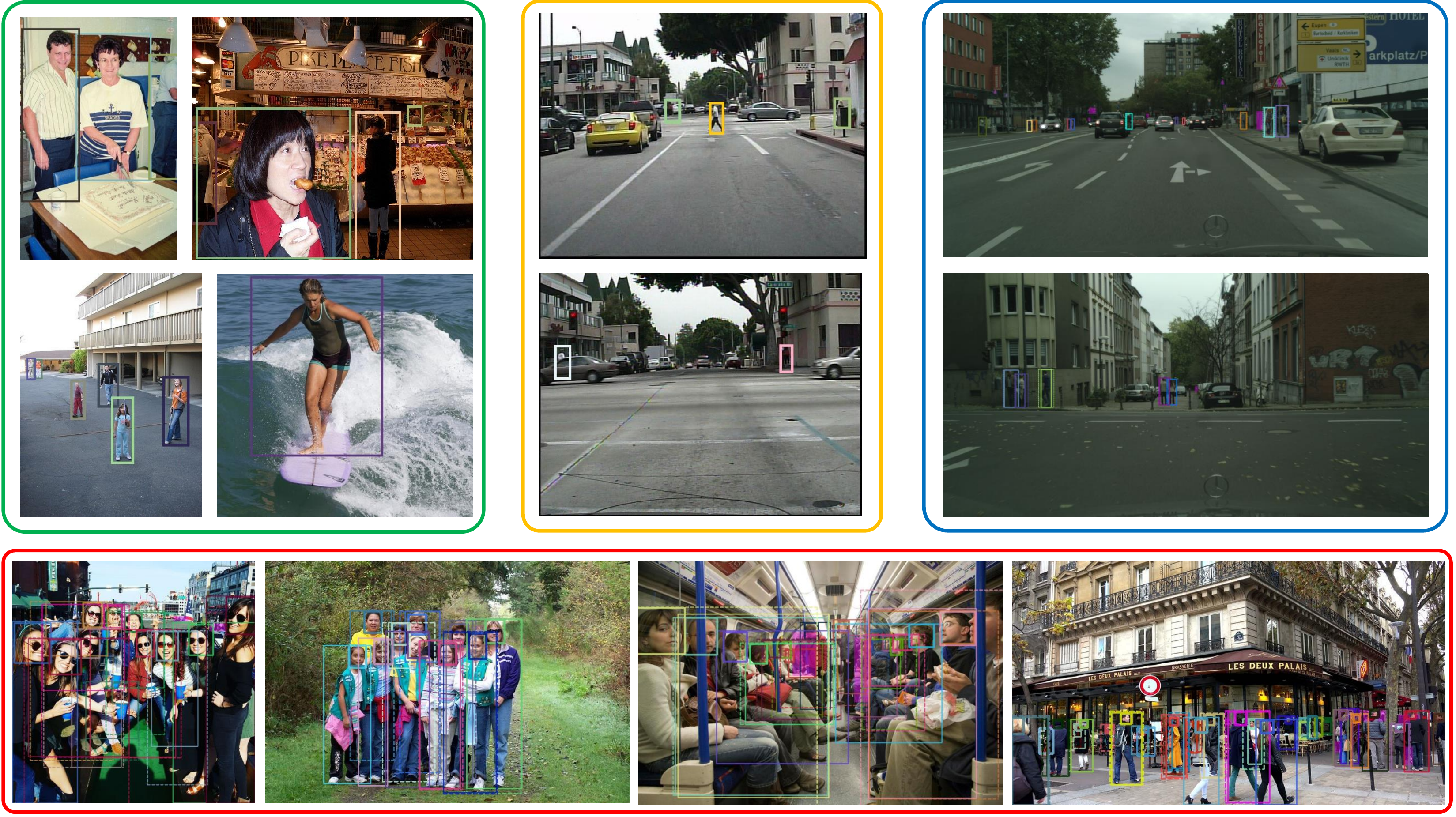}
\caption{Illustrative examples from different human dataset benchmarks. The images inside the green, yellow, blue boxes are from the COCO~\cite{lin2014microsoft}, Caltech~\cite{dollar2009pedestrian}, and CityPersons~\cite{zhang2017citypersons} datasets, respectively. The images from the second row inside the red box are from our CrowdHuman benchmark with full body, visible body, and head bounding box annotations for each person.}
\label{fig:dataset_overview}
\end{figure*}

Our goal is to push the boundary of human detection by specifically targeting the challenging crowd scenarios. We collect and annotate a rich dataset,  termed \emph{CrowdHuman}, with considerable amount of crowded pedestrians. CrowdHuman contains $15,000$, $4,370$ and $5,000$ images for training, validation, and testing respectively. The dataset is exhaustively annotated and contains diverse scenes. There are totally $470k$ individual persons in the train and validation subsets, and the average number of pedestrians per image reaches $22.6$. We also provide the visible region bounding-box annotation, and head region bounding-box annotation along with its full body annotation for each person. Fig.~\ref{fig:dataset_overview} shows examples in our dataset compared with those in other human detection datasets.

To summarize, we propose a new dataset called CrowdHuman with the following three contributions:
\begin{itemize}
    \item To the best of our knowledge, this is the first dataset which specifically targets to address the crowd issue in human detection task. More specifically, the average number of persons in an image is $22.6$ and the average of pairwise overlap between two human instances (larger than 0.5 IoU) is 2.4, both of which are much larger than the existing benchmarks like CityPersons, KITTI and Caltech. 
    \item The proposed CrowdHuman dataset provides annotations with three categories of bounding boxes: head bounding-box, human visible-region bounding-box, and human full-body bounding-box. Furthermore, these three categories of bounding-boxes are bound for each human instance.
    \item Experiments of cross-dataset generalization ability demonstrate our dataset can serve as a powerful pre-training dataset for many human detection tasks. A framework originally designed for general object detection without any specific modification provides state-of-the-art results on every previous benchmark including Caltech and CityPersons for pedestrian detection, COCOPerson for person detection, and Brainwash for head detection.
\end{itemize}


\section{Related Work}
\subsection{Human detection datasets.} 
Pioneer works of pedestrian detection datasets involve INRIA~\cite{dalal2005histograms}, TudBrussels~\cite{wojek2009multi}, and Daimler~\cite{enzweiler2009monocular}. These datasets have contributed to spurring interest and progress of human detection, However, as algorithm performance improves, these datasets are replaced by larger-scale datasets like Caltech-USA~\cite{dollar2009pedestrian} and KITTI~\cite{Geiger2012CVPR}. More recently, Zhang \etal~build a rich and diverse pedestrian detection dataset CityPersons~\cite{zhang2017citypersons} on top of CityScapes~\cite{cordts2016cityscapes} dataset. It is recorded by a car traversing various cities, contains dense pedestrians, and is annotated with high-quality bounding boxes. 

Despite the prevalence of these datasets, they all suffer a problem of from low density. Statistically, the Caltech-USA and KITTI datasets have less than one person per image, while the CityPersons has $\sim\negmedspace 6$ persons per image. In these datasets, the crowd scenes are significantly under-represented. Even worse, protocols of these datasets allow annotators to ignore and discard the regions with a large number of persons as exhaustively annotating crowd regions is incredibly difficult and time consuming. 


\begin{figure*}
\centering
\includegraphics[width=14cm]{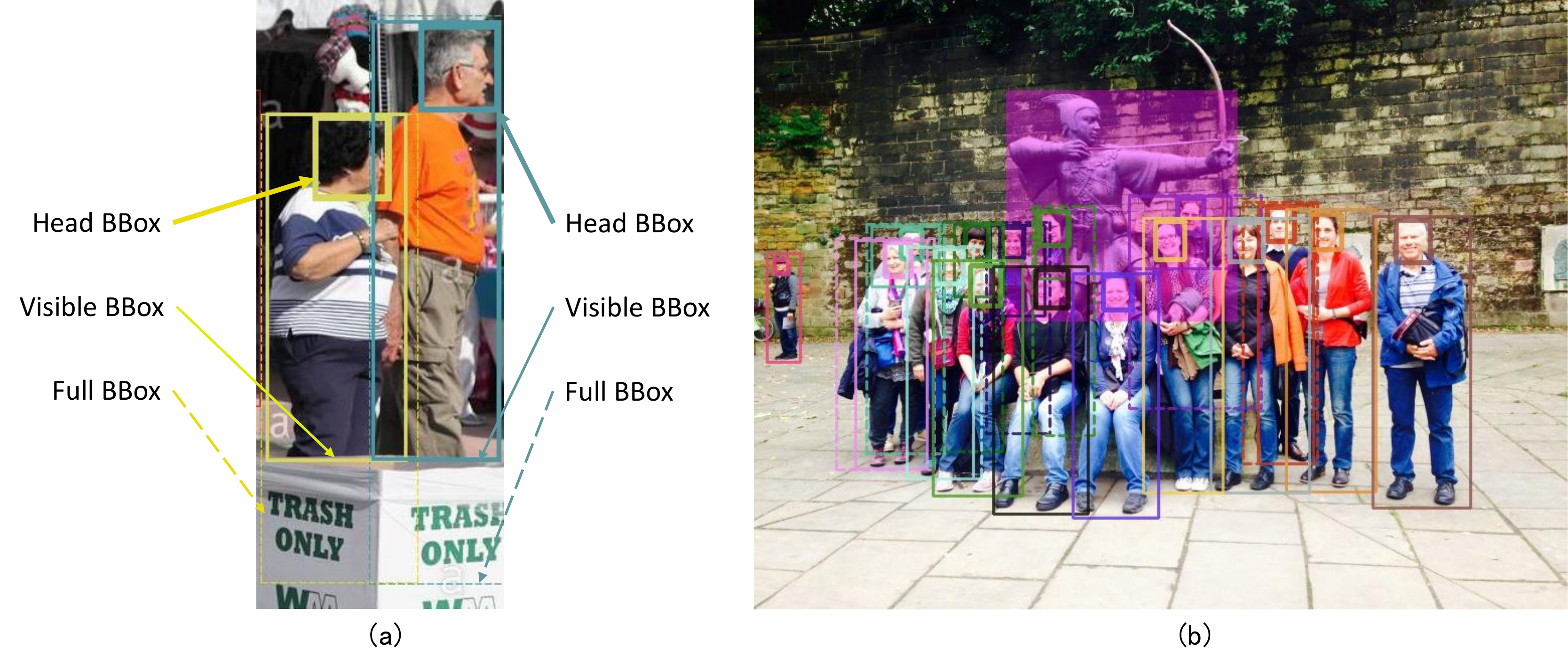}
\caption{(a) provides an illustrative example of our three kinds of annotations: Head Bounding-Box, Visible Bounding-Box, and Full Bounding-Box. (b) is an example image with our human annotations where magenta mask illustrates the ignored region.}
\label{fig:annotation_example}
\end{figure*}

\myparagraph{Human detection frameworks.} 
Traditional human detectors, such as ACF~\cite{dollar2014fast}, LDCF~\cite{nam2014local}, and Checkerboard~\cite{zhang2015filtered}, exploit various filters based on Integral Channel Features (IDF)~\cite{dollar2009integral} with sliding window strategy. 

Recently, the CNN-based detectors have become a predominating trend in the field of pedestrian detection. In~\cite{zhang2016faster}, self-learned features are extracted from deep neural networks and a boosted decision forest is used to detect pedestrians. Cai \etal~\cite{cai2016unified} propose an architecture which uses different levels of features to detect persons at various scales. Mao \etal~\cite{mao2017can} propose a multi-task network to further improve detection performance. Hosang \etal~\cite{Hosang_2017_CVPR} propose a learning method to improve the robustness of NMS. Part-based models are utilized in~\cite{ouyang2012discriminative,zhou2017multi} to alleviate occlusion problem. Repulsion loss is proposed to detect persons in crowd scenes~\cite{wang2018Repulsion}.

\section{CrowdHuman Dataset}
\label{sec:dataset}
In this section, we describe our CrowdHuman dataset including the collection process, annotation protocols, and informative statistics.
\subsection{Data Collection}
We would like our dataset to be diverse for real world scenarios. Thus, we crawl images from Google image search engine with $\sim\negmedspace 150$ keywords for query. Exemplary keywords include ``Pedestrians on the Fifth Avenue'', ``people crossing the roads'', ``students playing basketball'' and ``friends at a party''. These keywords cover more than $40$ different cities around the world, various activities (\eg, party, traveling, and sports), and numerous viewpoints (\eg, surveillance viewpoint and horizontal viewpoint). The number of images crawled from a keyword is limited to $500$ to make the distribution of images balanced. We crawl $\sim\negmedspace 60,000$ candidate images in total. The images with only a small number of persons, or with small overlaps between persons, are filtered. Finally, $\sim\negmedspace 25,000$ images are collected in the CrowdHuman dataset. We randomly select $15,000$, $4,370$ and $5,000$ images for training, validation, and testing, respectively.

\begin{table*}[!t]
\centering
\begin{tabular}{l|ccccc}
\hline
\hline
 & Caltech & KITTI & CityPersons & COCOPersons & CrowdHuman \\
\hline
Full BBox & $\checkmark$ & $\checkmark$ & $\checkmark^{\dagger}$ & $\times$ & $\checkmark$\\
Visible BBox & $\checkmark$ & $\times$  & $\checkmark$ & $\checkmark$ & $\checkmark$ \\
Head BBox & $\times$ & $\times$ & $\times$ & $\times$ & $\checkmark$ \\
\hline
\end{tabular}
\caption{\label{tab:annotation_comparison} Comparison of different annotation types for the popular human detection benchmarks.${}^{\dagger}:$ Aligned to a certain ratio.}
\end{table*}

\begin{table*}[!ht]
\centering
\begin{tabular}{l|ccccc}
\hline
\hline
 & Caltech & KITTI & CityPersons & COCOPersons & CrowdHuman \\
\hline
\# images & $42,782$ & $3,712$ & $2,975$ & ${\bf 64,115}$ & $15,000$ \\
\# persons & $13,674$ & $2,322$ & $19,238$ & $257,252$ & $\bf{339,565}$ \\
\# ignore regions & $50,363$ & $45$ & $6,768$ & $5,206$ & $\bf{99,227}$ \\
\# person/image & $0.32$ & $0.63$ & $6.47$ & $4.01$ & $\bf{22.64}$ \\
\# unique persons & $1,273$ & $< 2,322$ & $19,238$ & $257,252$ & $\bf{339,565}$ \\
\hline
\end{tabular}
\caption{\label{tab:dataset_statistics} Volume, density and diversity of different human detection datasets. For fair comparison, we only show the statistics of training subset.}
\end{table*}

\subsection{Image Annotation}
We annotate individual persons in the following steps. 
\begin{itemize}
    \item We annotate a \emph{full bounding box} of each individual exhaustively. If the individual is partly occluded, the annotator is required to complete the invisible part and draw a full bounding box. Different from the existing datasets like CityPersons, where the bounding boxes annotated are generated via drawing a line from top of the head and the middle of feet with a fixed aspect ratio (0.41), our annotation protocol is more flexible in real world scenarios which have various human poses. We also provide bounding boxes for human-like objects, \eg, statue, with a specific label. Following the metrics of~\cite{dollar2009pedestrian}, these bounding-boxes will be ignored during evaluation.
    \item We crop each annotated instance from the images, and send these cropped regions for annotators to draw a \emph{visible bounding box}.
    \item We further send the cropped regions to annotate a \emph{head bounding box}. All the annotations are double-checked by at least one different annotator to ensure the annotation quality.
\end{itemize}
Fig.~\ref{fig:annotation_example} shows the three kinds of bounding boxes associated with an individual person as well as an example of annotated image. 

We compare our CrowdHuman dataset with previous datasets in terms of annotation types in Table~\ref{tab:annotation_comparison}. Besides from the popular pedestrian detection datasets, we also include the COCO~\cite{lin2014microsoft} dataset with only a ``person" class. Compared with CrowdHuman, which provides various types of annotations, Caltech and CityPersons have only normalized full bounding boxes and visible boxes, KITTI has only full bounding boxes, and COCOPersons has only visible bounding boxes. More importantly, none of them has head bounding boxes associated with each individual person, which may serve as a possible means to address the crowd occlusion problem.

\subsection{Dataset Statistics}
\label{sec:dataset_statistics}
\paragraph{Dataset Size.}
The volume of the CrowdHuman training subset is illustrated in the first three lines of Table~\ref{tab:dataset_statistics}. In a total of $15,000$ images, there are $\sim\negmedspace 340k$ person and $\sim\negmedspace 99k$ ignore region annotations in the CrowdHuman training subset. The number is more than 10x boosted compared with previous challenging pedestrian detection dataset like CityPersons. The total number of persons is also noticeably larger than the others. 

\paragraph{Density.}
In terms of density, on average there are $\sim\negmedspace 22.6$ persons per image in CrowdHuman dataset, as shown in the fourth line of Table~\ref{tab:dataset_statistics}. We also report the density from the existing datasets in Table~\ref{tab:density}. Obviously, CrowdHuman dataset is of much higher crowdness compared with all previous datasets. Caltech and KITTI suffer from extremely low-density, for that on average there is only $\sim\negmedspace 1$ person per image. The number in CityPersons reaches $\sim\negmedspace 7$, a significant boost while still not dense enough. As for COCOPersons, although its volume is relatively large, it is insufficient to serve as a ideal benchmark for the challenging crowd scenes. Thanks to the pre-filtering and annotation protocol of our dataset, CrowdHuman can reach a much better density.

\paragraph{Diversity.}
Diversity is an important factor of a dataset. 
COCOPersons and CrowdHuman contain people in unlimited poses in a wide range of domains, while Caltech, KITTI and CityPersons are all recorded by a car traversing on streets. 
The number of identical persons is also critical. As reported in the fifth line in Table~\ref{tab:dataset_statistics}, this number amounts to $\sim\negmedspace 33k$ in CrowdHuman while images in Caltech and KITTI are \emph{not} sparsely sampled, resulting in less amount of identical persons.

\begin{figure*}
\centering
\includegraphics[height=7cm]{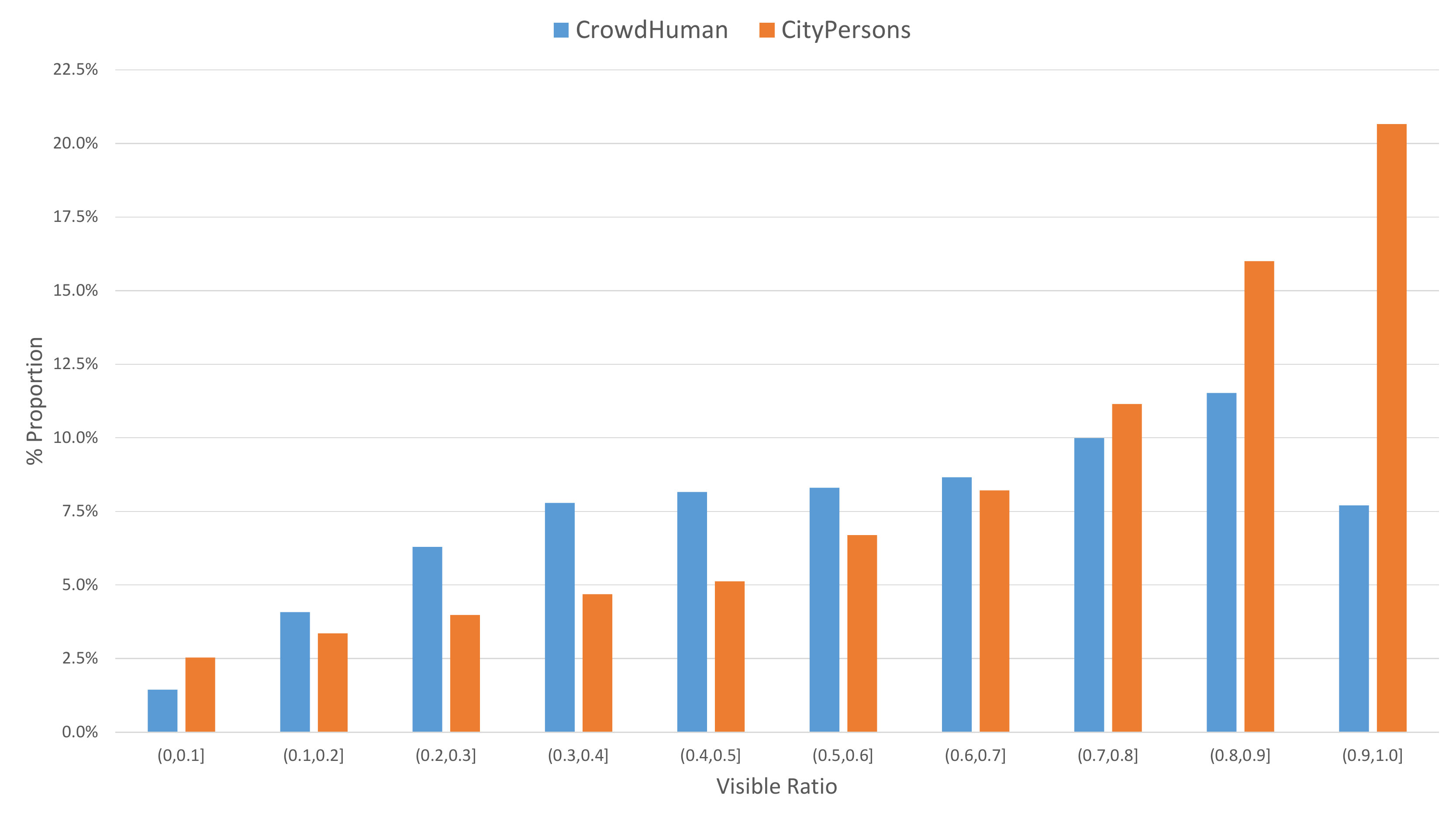}
\caption{Comparison of the visible ratio between our CrowdHuman and CityPersons dataset. Visible Ratio is defined as the ratio of visible bounding box to the full bounding box.}
\label{fig:visRatio}
\end{figure*}

\paragraph{Occlusion.}
To better analyze the distribution of occlusion levels, we divide the dataset into the ``bare'' subset ($\mathrm{occlusion} \leq 30\%$), the ``partial'' subset ($30\% < \mathrm{occlusion }\leq 70\%$), and the ``heavy'' subset ($\mathrm{occlusion} > 70\%$). In Fig.~\ref{fig:visRatio}, we compare the distribution of persons at different occlusion levels for CityPersons\footnote{The statistics is computed without group people}. The bare subset and partial subset in CityPersons constitute $46.79\%$ and $24.19\%$ of entire dataset respectively, while the ratios for CrowdHuman are $29.89\%$ and $32.13\%$. The occlusion levels are more balanced in CrowdHuman, in contrary to those in CityPersons, which have more persons with low occlusion.

We also provide statistics on pair-wise occlusion. For each image, We count the number of person pairs with different intersection over union (IoU) threshold. The results are shown in Table~\ref{tab:PairOverlap}. In average, few person pairs with an IoU threshold of $0.3$ are included in Caltech, KITTI or COCOPersons. For CityPersons dataset, the number is less than one pair per image. However, the number is $9$ for CrowdHuman. Moreover, There are averagely $2.4$ pairs whose IoU is greater than $0.5$ in the CrowdHuman dataset. We further count the occlusion levels for triples of persons. As shown in Table~\ref{tab:highorderOverlap}, such cases can be hardly found in previous datasets, while they are well-represented in CrowdHuman.

\begin{table*}
\begin{center}
\begin{tabular}{c|cc|cc|cc|cc|cc}
{person/img $\ge$} & \multicolumn{2}{c}{Caltech} & \multicolumn{2}{c}{KITTI} & \multicolumn{2}{c}{CityPersons} & \multicolumn{2}{c}{COCOPersons} & \multicolumn{2}{c}{CrowdHuman} \\
\hline
{1} & {7839} & {18.3\%} & {969} & {26.1\%} & {2482} & {83.4\%} & {64115} & {100.0\%} & {15000} & {100.0\%} \\
{2} & {3257} & {7.6\%} & {370} & {10.0\%} & {2082} & {70.0\%} & {39283} & {61.3\%} & {15000} & {100.0\%} \\
{3} & {1265} & {3.0\%} & {273} & {7.4\%} & {1741} & {58.5\%} & {28553} & {44.5\%} & {14996} & {100.0\%} \\
{5} & {282} & {0.7\%} & {164} & {4.4\%} & {1225} & {41.2\%} & {18775} & {29.3\%} & {14220} & {94.8\%} \\
{10} & {36} & {0.1\%} & {19} & {0.5\%} & {610} & {20.5\%} & {9604} & {15.0\%} & {10844} & {72.3\%} \\
{20} & {0} & {0.0\%} & {0} & {0.0\%} & {227} & {7.6\%} & {0} & {0.0\%} & {5907} & {39.4\%} \\
{30} & {0} & {0.0\%} & {0} & {0.0\%} & {94} & {3.2\%} & {0} & {0.0\%} & {3294} & {21.9\%} \\
\end{tabular}
\caption{Comparison of the human density against the widely used human detection dataset. The first column refers to the number of human instances in the image. } \label{tab:density}
\end{center}
\end{table*}

\begin{table}
\begin{center}
\begin{tabular}{c|cccc}
{pair/img} & {Cal} & {City} & {COCO} & {CrowdHuman} \\
\hline
{iou$>$0.3} & {0.06} & {0.96} & {0.13} & {9.02} \\
{iou$>$0.4} & {0.03} & {0.58} & {0.05} & {4.89} \\
{iou$>$0.5} & {0.02} & {0.32} & {0.02} & {2.40} \\
{iou$>$0.6} & {0.01} & {0.17} & {0.01} & {1.01} \\
{iou$>$0.7} & {0.00} & {0.08} & {0.00} & {0.33} \\
{iou$>$0.8} & {0.00} & {0.02} & {0.00} & {0.07} \\
{iou$>$0.9} & {0.00} & {0.00} & {0.00} & {0.01} \\
\end{tabular}
\caption{Comparison of pair-wise overlap between two human instances.} \label{tab:PairOverlap}
\end{center}

\end{table}

\begin{table}
\begin{center}
\begin{tabular}{c|cccc}
{pair/img} & \multicolumn{1}{c}{Cal} & \multicolumn{1}{c}{City} & \multicolumn{1}{c}{COCO} & \multicolumn{1}{c}{CrowdHuman} \\
\hline
{iou$>$0.1} & {0.02} & {0.30} & {0.02} & {8.70} \\
{iou$>$0.2} & {0.00} & {0.11} & {0.00} & {2.09} \\
{iou$>$0.3} & {0.00} & {0.04} & {0.00} & {0.51} \\
{iou$>$0.4} & {0.00} & {0.01} & {0.00} & {0.12} \\
{iou$>$0.5} & {0.00} & {0.00} & {0.00} & {0.03} \\
\end{tabular}
\end{center}
\caption{Comparison of high-order overlaps among three human instances.} \label{tab:highorderOverlap}
\end{table}

\section{Experiments}

In this section, we will first discuss the experiments on our CrowdHuman dataset, including full body detection, visible body detection and head detection. Meanwhile, the generalization ability of our CrowdHuman dataset will be evaluated on standard pedestrian benchmarks like Caltech and CityPersons, person detection benchmark on COCOPersons, and head detection benchmark on Brainwash dataset. We use FPN~\cite{lin2017feature} and RetinaNet~\cite{lin2017focal} as two baseline detectors to represent the two-stage algorithms and one-stage algorithms, respectively. 

\subsection{Baseline Detectors}
Our baseline detectors are Faster R-CNN~\cite{ren2015faster} and RetinaNet~\cite{lin2017focal}, both based on the Feature Pyramid Network (FPN)~\cite{lin2017feature} with a ResNet-50~\cite{he2016deep} back-bone network. Faster R-CNN and RetinaNet are both proposed for general object detection, and they have dominated the field of object detection in recent years. 


\subsection{Evaluation Metric}
The training and validation subsets of CrowdHuman can be downloaded from our website. In the following experiments, our algorithms are trained based on CrowdHuman train subset and the results are evaluated in the validation subset.  An online evaluation server will help to evaluate the performance of the testing subset and a leaderboard will be maintained. The annotations of testing subset will not be made publicly available.

We follow the evaluation metric used for Caltech~\cite{dollar2009pedestrian}, denoted as mMR, which is the average log miss rate over false positives per-image ranging in $\left[ 10^{-2}, 10^0\right]$. mMR is a good indicator for the algorithms applied in the real world applications. Results on ignored regions will not considered in the evaluation. Besides, Average Precision (AP) and recall of the algorithms are included for reference.

\subsection{Implementation Details}
We use the same setting of anchor scales as~\cite{lin2017feature} and~\cite{lin2017focal}. For all the experiments related to full body detection, we modify the height v.s. width ratios of anchors as $\left\{ 1:1, 1.5:1, 2:1, 2.5:1, 3:1 \right\}$ in consideration of  the human body shape. While for visible body detection and human head detection, the ratios are set to $\left\{ 1:2, 1:1, 2:1 \right\}$, in comparison with the original papers. The input image sizes of Caltech and CityPersons are set to $2\times$ and $1\times$ of the original images according to~\cite{zhang2017citypersons}. As the images of CrowdHuman and MSCOCO are both collected from the Internet with various sizes, we resize the input so that their short edge is at $800$ pixels while the long edge should be no more than $1400$ pixels at the same time. The input sizes of Brainwash is set as $640\times480$.

We train all datasets with $600k$ and $750k$ iterations for FPN and RetinaNet, respectively. The base learning rate is set to $0.02$ and decreased by a factor of $10$ after $150k$ and $450k$ for FPN, and $180k$ and $560k$ for RetinaNet. The Stochastic Gradient Descent (SGD) solver is adopted to optimize the networks on $8$ GPUs. A mini-batch involves $2$ images per GPU, except for CityPersons where a mini-batch involves only $1$ image due to the physical limitation of GPU memory. Weight decay and momentum are set to $0.0001$ and $0.9$. We do not finetune the batch normalization~\cite{ioffe2015batch} layers. Multi-scale training/testing are not applied to ensure fair comparisons.

\subsection{Detection results on CrowdHuman}
\label{sec:expCrowdhuman}

\begin{figure*}
\centering
\includegraphics[width=12cm]{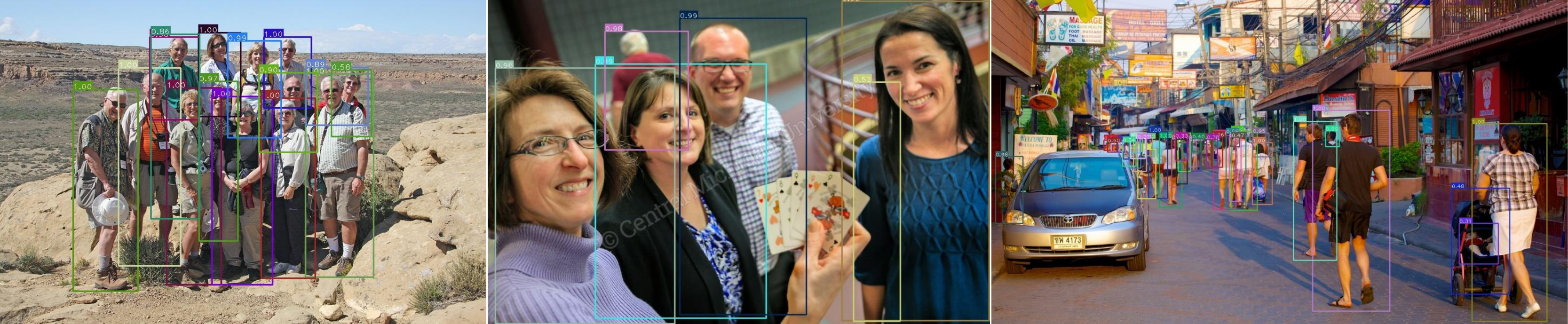}
\caption{Qualitative results for the visible body detection of FPN based on CrowdHuman dataset.}
\label{fig:result_fpn_vis}
\end{figure*}

\myparagraph{Visible Body Detection} As the human have different poses and occlusion conditions, the visible regions may be quite different for each individual person, which brings many difficulties to human detection. Table~\ref{table:evalvis} illustrates the results for the visible part detection based on FPN and RetinaNet. FPN outperforms RetinaNet in this case. According to Table~\ref{table:evalvis}, the proposed CrowdHuman dataset is a challenging benchmark, especially for the state-of-the-art human detection algorithms. The illustrative examples of visible body detection based on FPN are shown in Fig.~\ref{fig:result_fpn_vis}.

\setlength{\tabcolsep}{4pt}
\begin{table}
\begin{center}
\caption{Evaluation of visible body detection on CrowdHuman benchmark.}
\label{table:evalvis}
\begin{tabular}{cccc}
\hline\noalign{\smallskip}
{} & Recall & AP & mMR\\
\noalign{\smallskip}
\hline
\noalign{\smallskip}
FPN~\cite{lin2017feature}  & {91.51} & {85.60} & {\textbf{55.94}}\\
RetinaNet~\cite{lin2017focal} & {90.96} & {77.19} & {65.47}\\
\hline
\end{tabular}
\end{center}
\end{table}
\setlength{\tabcolsep}{1.4pt}

\myparagraph{Full Body Detection} 
Detecting full body regions is more difficult than detecting the visible part as the detectors should predict the occluded boundaries of the full body. To make matters worse, the ground-truth annotation might be suffered from high variance caused by different decision-makings by different annotators.

\begin{figure*}
\centering
\includegraphics[width=12cm]{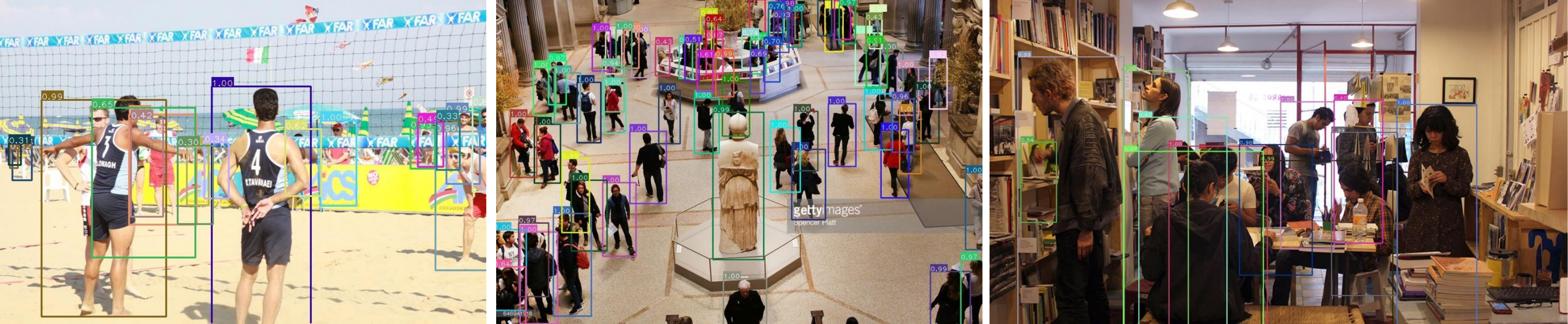}
\caption{Qualitative results for the full body detection of FPN based on CrowdHuman dataset.}
\label{fig:result_fpn_full}
\end{figure*}

Different from the visible part detection, the aspect ratios of the anchors for the full body detection are set as $[1.0, 1.5, 2.0, 2.5, 3.0]$ to make the detector tend to predict the slim and tall bounding boxes. Another important thing is that the RoIs are not clipped into the limitation of the image boundaries, as there are many full body bounding boxes extended out of images. The results are shown in Table~\ref{table:evalfull} and the illustrative examples of FPN are shown in Fig.~\ref{fig:result_fpn_full}. Similar to the Visible body detection, FPN has a significant gain over RetinaNet.

\setlength{\tabcolsep}{4pt}
\begin{table}
\begin{center}
\caption{Evaluation of full body detection on CrowdHuman benchmark.}
\label{table:evalfull}
\begin{tabular}{cccc}
\hline\noalign{\smallskip}
{} & Recall & AP & mMR\\
\noalign{\smallskip}
\hline
\noalign{\smallskip}
FPN~\cite{lin2017feature} & {90.24} & {84.95} & {\textbf{50.42}}\\
RetinaNet~\cite{lin2017focal} & {93.80} & {80.83} & {63.33}\\
\hline
\hline
FPN on Caltech & {99.76}  & {89.95} & {10.08}\\
FPN on CityPersons & {97.97} & {94.35} & {14.81}\\
\hline
\end{tabular}
\end{center}
\end{table}
\setlength{\tabcolsep}{1.4pt}

In Table~\ref{table:evalfull}, we also report the FPN pedestrian detection results\footnote{The results are evaluated on the standard reasonable set} on Caltech, i.e., 10.08 mMR, and CityPersons, i.e., 14.81 mMR. It shows that our CrowdHuman dataset is much challenging than the standard pedestrian detection benchmarks based on the detection performance.

\begin{figure*}
\centering
\includegraphics[width=12cm]{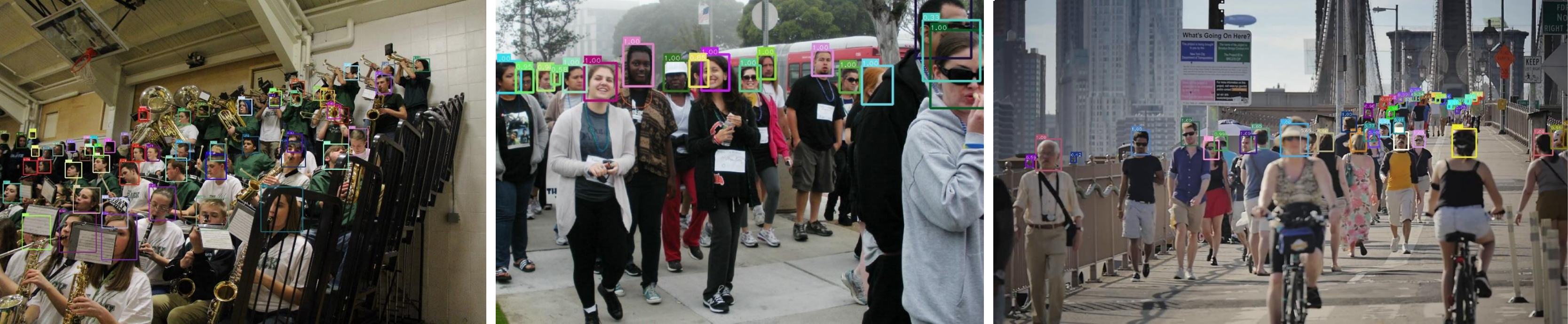}
\caption{Qualitative results for the head detection of FPN based on CrowdHuman dataset.}
\label{fig:result_fpn_head}
\end{figure*}

\myparagraph{Head Detection} Head is one of the most obvious parts of a whole body. Head detection is widely used in the practical applications such as people number counting, face detection and tracking. We compare the results of FPN and RetinaNet as shown in Table~\ref{table:evalHead}. The illustrative examples of head detection on CrowdHuman by FPN detector are shown in Fig.~\ref{fig:result_fpn_head}.

\setlength{\tabcolsep}{4pt}
\begin{table}
\begin{center}
\caption{Evaluation of Head detection on CrowdHuman benchmark.}
\label{table:evalHead}
\begin{tabular}{cccc}
\hline\noalign{\smallskip}
{} & Recall & AP & mMR\\
\noalign{\smallskip}
\hline
\noalign{\smallskip}
FPN~\cite{lin2017feature}  & {81.10} & {77.95} & {\textbf{52.06}}\\
RetinaNet~\cite{lin2017focal} & {78.43} & {71.36} & {60.64}\\
\hline
\end{tabular}
\end{center}
\end{table}
\setlength{\tabcolsep}{1.4pt}

\subsection{Cross-dataset Evaluation}\label{exp:crossdataset}

As shown in Section~\ref{sec:dataset}, the size of CrowdHuman dataset is obviously larger than the existing benchmarks, like Caltech and CityPersons. In this section, we evaluate that the generalization ability of our CrowdHuman dataset. More specifically, we first train the model on our CrowdHuman dataset and then finetune it on the visible body detection benchmarks like COCOPersons~\cite{lin2014microsoft}, full body detection benchmarks like Caltech~\cite{dollar2009pedestrian} and CityPersons~\cite{zhang2017citypersons}, and head detection benchmarks like Brainwash~\cite{stewart2016end}. As reported in Section~\ref{sec:expCrowdhuman}, FPN is superior to RetinaNet in all three cases. Therefore, in the following experiments, we adopt FPN as our baseline detector.

\myparagraph{COCOPersons} COCOPersons is a subset of MSCOCO from the images with groundtruth bounding box of ``person''. The other 79 classes are ignored in our evaluation. After the filtering process, there are 64115 images from the trainval minus minival for training, and the other 2639 images from minival for validation. All the persons in COCOPersons are annotated as the visible body with different type of human poses. The results are illustrated in Table~\ref{table:cocoperson}. Based on the pretraining of our CrowdHuman dataset, our algorithm has superior performance on the COCOPersons benchmark against the one without CrowdHuman pretraining. 

\setlength{\tabcolsep}{4pt}
\begin{table}
\begin{center}
\caption{Experimental results on COCOPersons.}
\label{table:cocoperson}
\begin{tabular}{cccc}
\hline\noalign{\smallskip}
{Train-set} & Recall & AP & mMR\\
\noalign{\smallskip}
\hline
\noalign{\smallskip}
COCOPersons & {95.57}  & {83.83} & {41.89}\\
Crowd$\Rightarrow$COCO & {95.87}  & {85.02} & {$\mathbf{39.79}$}\\
\hline
\end{tabular}
\end{center}
\end{table}
\setlength{\tabcolsep}{1.4pt}

\myparagraph{Caltech and CityPersons} Caltech and CityPersons are widely used benchmarks for pedestrian detection, both of them are usually adopted to evaluate full body detection algorithms. 
We use the reasonable set for Caltech dataset where the object size is larger than 50 pixels.  
Table~\ref{table:evalCaltech} and Table~\ref{table:evalCity} show the results on Caltech and CityPersons, respectively. We compare the algorithms in the first part of the tables with:
\begin{itemize}
    \item FPN trained on the Caltech
    \item FPN trained on CityPersons
    \item FPN trained on CrowdHuman
    \item FPN model pretrained on CrowdHuman and then finetuned on the corresponding target training set
\end{itemize}
Also, state-of-art algorithms on Caltech and CityPersons are reported in the second part of tables as well.
To summarize, the results illustrated in Table~\ref{table:evalCaltech} and Table~\ref{table:evalCity}  demonstrate that our CrowdHuman dataset can serve as an effective pretraining dataset for pedestrian detection task on Caltech and CityPersons~\footnote{The evaluation is based on $1\times$ scale.} for full body detection.

\begin{table*}
\parbox{.5\linewidth}{
\centering
\caption{Experimental results on Caltech dataset.}
\label{table:evalCaltech}
\begin{tabular}{lccc}
\hline
Train-set & Recall & AP & mMR\\
\hline
Caltech & {99.76}  & {89.95} & {10.08}\\
CityPersons & {99.05} & {85.81} & {14.69}\\
CrowdHuman & {99.88} & {90.58} & {8.81}\\
Crowd$\Rightarrow$Calt & {99.88} & {95.69} & {$\bf{3.46}$}\\
\hline
\hline
CityPersons$\Rightarrow$Calt~\cite{zhang2017citypersons} & - & - & {5.1}\\
Repulsion~\cite{wang2018Repulsion} & - & - & {4.0}\\
\cite{mao2017can} & - & - & {5.5}\\
\hline
\end{tabular}
}
\hfill
\parbox{.5\linewidth}{
\centering
\caption{Experimental reslts on CityPersons.}
\label{table:evalCity}
\begin{tabular}{lcccc}
\hline\noalign{\smallskip}
Train-set & Recall & AP & mMR\\
\noalign{\smallskip}
\hline
\noalign{\smallskip}
Caltech  & {87.21}  & {65.87} & {45.52}\\
CityPersons & {97.97} & {94.35} & {14.81}\\
CrowdHuman & {98.73} & {98.10} & {21.18}\\
Crowd$\Rightarrow$City & {97.78} & {95.58} & {$\bf{10.67}$}\\
\hline
\hline
CityPersons~\cite{zhang2017citypersons} & - & - & {14.8}\\
Repulsion~\cite{wang2018Repulsion} & - & - & {13.2}\\
\hline
\end{tabular}
}
\end{table*}

\myparagraph{Brainwash} Brainwash~\cite{stewart2016end} is a head detection dataset whose images are extracted from the video footage at every 100 seconds. Following the step of~\cite{stewart2016end}, the training set has 10,917 images with 82,906 instances and the validation set has 500 images with 3318 instances. Similar to visible body detection and full body detection, Brainwash dataset is evaluated to validate the generalization ability of our CrowdHuman dataset for head detection.

Table~\ref{table:brainwash} shows the results of head detection task on Brainwash dataset. By using the FPN as the head detector, the performance is already much better than the state-of-art in~\cite{stewart2016end}. On top of that, pretraining on the CrowdHuman dataset further boost the result by 2.5\% of mMR, which validates the generalization ability of our CrowdHuman dataset for head detection.

\setlength{\tabcolsep}{4pt}
\begin{table}
\begin{center}
\caption{Experimental results on Brainwash.}
\label{table:brainwash}
\begin{tabular}{cccc}
\hline\noalign{\smallskip}
{Train-set} & Recall & AP & mMR\\
\noalign{\smallskip}
\hline
\noalign{\smallskip}
Brainwash & {98.52} & {95.74} & {19.77}\\
Crowd$\Rightarrow$Brain & {98.66}  & {96.15} & {$\mathbf{17.24}$}\\
\hline
\hline
\cite{stewart2016end} & - & {78.0} & -\\
\hline
\end{tabular}
\end{center}
\end{table}
\setlength{\tabcolsep}{1.4pt}

\section{Conclusion}
In this paper, we present a new human detection benchmark designed to address the crowd problem. There are three contributions of our proposed CrowdHuman dataset. Firstly, compared with the existing human detection benchmark, the proposed dataset is larger-scale with much higher crowdness. Secondly, the full body bounding box, the visible bounding box, and the head bounding box are annotated for each human instance. The rich annotations enables a lot of potential visual algorithms and applications. Last but not least, our CrowdHuman dataset can serve as a powerful pretraining dataset. State-of-the-art results have been reported on benchmarks of pedestrian detection benchmarks like Caltech and CityPersons, and Head detection benchmark like Brainwash. The dataset as well as the code and models discussed in the paper will be released~\footnote{\url{https://sshao0516.github.io/CrowdHuman/}}.

{\small
\bibliographystyle{ieee}
\bibliography{egbib}
}

\end{document}